%% file: main.tex
\documentclass[letterpaper]{article} 
\usepackage{aaai24}
\usepackage{times}  
\usepackage{helvet}  
\usepackage{courier}  
\usepackage[hyphens]{url}  
\usepackage{graphicx} 
\usepackage{natbib}  
\usepackage{caption} 
\usepackage{booktabs}
\usepackage{subfigure}
\usepackage{algorithm}
\usepackage{algorithmic}
\usepackage{amsmath}
\usepackage{newfloat}
\usepackage{listings}
\DeclareCaptionStyle{ruled}{labelfont=normalfont,labelsep=colon,strut=off} 
\floatstyle{ruled}
\newfloat{listing}{tb}{lst}{}
\floatname{listing}{Listing}

\nocopyright 

\title{Delving into Multimodal Prompting for Fine-grained Visual Classification}

\author{
	Xin Jiang\textsuperscript{\rm 1}\thanks{Equal contribution.},
	Hao Tang\textsuperscript{\rm 1}\footnotemark[1],
	Junyao Gao\textsuperscript{\rm 2},
	Xiaoyu Du\textsuperscript{\rm 1},
        Shengfeng He\textsuperscript{\rm 3},
	Zechao Li\textsuperscript{\rm 1}\thanks{Corresponding author.}
}

\affiliations{
    \textsuperscript{\rm 1}Nanjing University of Science and Technology, China\\
    \textsuperscript{\rm 2}Tongji University, China\\
    \textsuperscript{\rm 3}Singapore Management University, Singapore\\
    \{xinjiang, tanghao0918, duxy, zechao.li\}@njust.edu.cn,  junyaogao@tongji.edu.cn, shengfenghe7@gmail.com
}

\usepackage{bibentry}
\usepackage{multirow}
\usepackage{makecell}
\usepackage{amssymb}
\usepackage{float}

\begin{document}
\maketitle
\input{sections/abstract}
\input{sections/introduction}
\input{sections/relatework}
\input{sections/preliminary}
\input{sections/Methodology}
\input{sections/Experiments}

\input{sections/Conlusion}
\section{Acknowledgements}
This work was partially supported by the National Key Research and Development Program of China under Grant 2022ZD0118802, the National Natural Science Foundation of China (Grant No. U20B2064 and 62172226), Singapore MOE Tier 1 Funds (MSS23C002), AISG Research Grant (No. AISG3-GV-2023-011),  and the Postgraduate Research \& Practice Innovation Program of Jiangsu Province (KYCX230491).

\bibliography{aaai24}

\end{document}

%% file: sections/abstract.tex
\section{Abstract}
Fine-grained visual classification (FGVC) involves categorizing fine subdivisions within a broader category, which poses challenges due to subtle inter-class discrepancies and large intra-class variations. However, prevailing approaches primarily focus on uni-modal visual concepts. Recent advancements in pre-trained vision-language models have demonstrated remarkable performance in various high-level vision tasks, yet the applicability of such models to FGVC tasks remains uncertain. In this paper, we aim to fully exploit the capabilities of cross-modal description to tackle FGVC tasks and propose a novel multimodal prompting solution, denoted as \textsc{MP-FGVC}, based on the contrastive language-image pertaining (CLIP) model. Our \textsc{MP-FGVC} comprises a multimodal prompts scheme and a multimodal adaptation scheme. The former includes Subcategory-specific Vision Prompt (\textsc{SsVP}) and Discrepancy-aware Text Prompt (\textsc{DaTP}), which explicitly highlights the subcategory-specific discrepancies from the perspectives of both vision and language. The latter aligns the vision and text prompting elements in a common semantic space, facilitating cross-modal collaborative reasoning through a Vision-Language Fusion Module (\textsc{VLFM}) for further improvement on FGVC. Moreover, we tailor a two-stage optimization strategy for \textsc{MP-FGVC} to fully leverage the pre-trained CLIP model and expedite efficient adaptation for FGVC. Extensive experiments conducted on four FGVC datasets demonstrate the effectiveness of our \textsc{MP-FGVC}.

%% file: sections/introduction.tex
\section{Introduction}
Fine-grained visual classification~(FGVC) is a significant task in the field of computer vision that aims to categorize distinct subcategories within a given supercategory. Unlike traditional classification tasks, FGVC faces the challenges of subtle inter-class variations caused by similar subordinate categories, as well as large intra-class variations due to factors such as location, scale, and deformation.  
Recognizing these subtle differences between subcategories, such as distinguishing bird subcategories based on characteristics like eye shape, flipper appearance, and coloration of the tail region, is a key aspect of human visual perception. To address these challenges, deep learning methods~\cite{YuZZZY18, mrdmn,GuF23, LYYL023, GuFHL23} have been proposed for FGVC, falling into two main categories: feature-encoding methods and part-localization methods. Part-localization methods, in particular, have garnered significant interest due to their consistency with human recognition intuition, as fine-grained subcategory-specific discrepancies~\cite{dong2023erasing} often reside in the unique properties of object parts.

Although CNN-based models have achieved satisfactory performance for FGVC in a weakly-supervised manner, a prominent drawback persists: CNNs tend to focus solely on a small, irrelevant region within an image, resulting in the captured subcategory-specific discrepancies that lack robustness and discriminability.
Recently, vision transformers like ViT~\cite{vit}, have recorded remarkable performance
in the image classification tasks and are actively being employed in FGVC~\cite{transfg,ielt}. Compared to CNNs, ViT’s patch-by-patch processing lends itself well to FGVC as each image patch can be regarded as a local part. Additionally, the inherent self-attention mechanism in ViT enables the modeling of long-range dependencies in the entire image.
However, due to the relatively limited size of FGVC datasets, these CNN-based and ViT-based methods often exhibit erratic behavior during optimization and heavily rely on visual representations from a predefined closed set during inference, which hampers the full exploitation of these models' potential for FGVC.

\begin{figure*}[t!]
    \centering
    \includegraphics[width=0.9\textwidth]{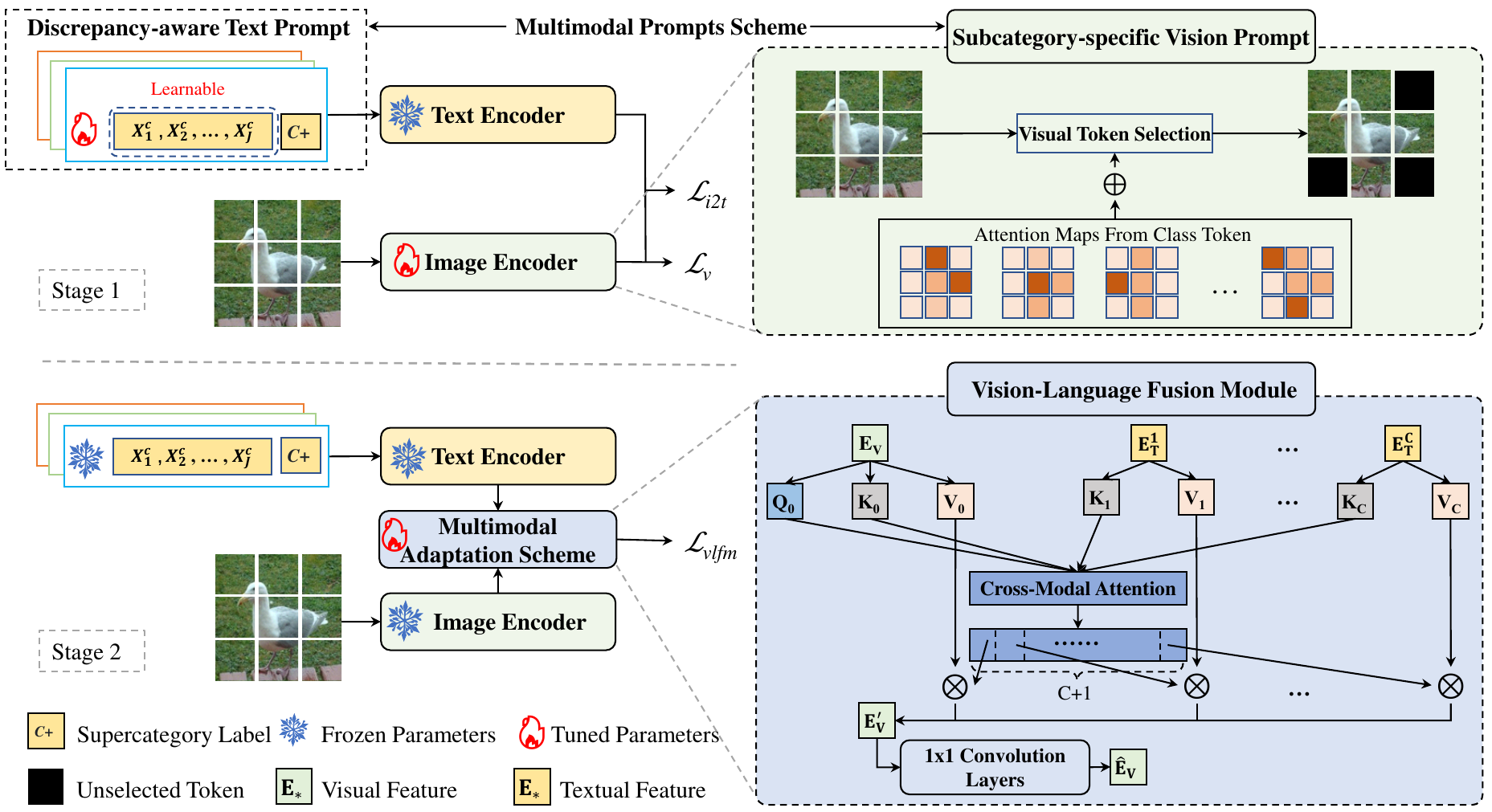}
    \caption{\small The pipeline of the proposed \textsc{MP-FGVC}, which consists of a multimodal prompts scheme and a multimodal adaptation scheme.}
    \label{fig:overview}\vspace{-4mm}
\end{figure*}

The human visual perception system possesses a unique mechanism of top-down \textbf{cognitive penetrability}~\cite{maier2019no} in which prior linguistic knowledge~\cite{li2023knowledge}, such as appearance descriptions, is instinctively used to adaptively adjust ongoing visual perceptual processing to stimulus features relevant to object categories, thereby facilitating the recognition of fine-grained objects. Recently, large-scale vision-language (VL) models have been developed to connect visual representation with high-level language descriptions, showing potential for adaptation to various vision tasks.
Thus, it raises the question of whether the cross-modal description ability in the VL models can be effectively utilized for FGVC tasks. Recent studies~\cite{abs-2203-17274, JiaTCCBHL22} have already confirmed the possibility of utilizing the knowledge from the CLIP model via prompt techniques~\cite{coop,cocoop} for classification tasks. 
However, existing prompt strategies are primarily designed to capture high-level category semantics rather than the more detailed subcategory-level discrepancies. Furthermore, the lack of specific annotations in existing FGVC datasets hampers the widespread adoption of VL models in FGVC.
Fortunately, recent works~\cite{li2023clip,pleor} have introduced large-scale contrastive language-image pretraining (CLIP)~\cite{clip} model to the domain of fine-grained understanding. However, these methods aim to associate a broader range of visual concepts, including both pre-defined and unknown categories~\cite{TangLPT20}, with their textual descriptions~\cite{yan2023image}, addressing the problem of fine-grained image retrieval in open-set scenarios. Therefore, it is worth investigating how to effectively make the pre-trained CLIP model sensitive to subcategory-specific discrepancies according to the semantic prompts to enhance the performance of FGVC tasks in close-set scenarios.

To this end, leveraging the capabilities of the recently introduced CLIP model, we present a multimodal prompting solution, called~\textsc{MP-FGVC}, for FGVC. The primary objective of \textsc{MP-FGVC} is to align the FGVC task with the classification task performed by the pre-trained CLIP model. Fig.~\ref{fig:overview} illustrates the proposed \textsc{MP-FGVC} framework, which consists of a multimodal prompts scheme and a multimodal adaptation scheme. 
The multimodal prompts scheme, comprising Subcategory-specific Vision Prompt~(\textsc{SsVP}) and Discrepancy-aware Text Prompt~(\textsc{DaTP}), explicitly emphasizes the subcategory-specific discrepancies from the input perspective. 
In particular, \textsc{SsVP} identifies subcategory-specific discrepancies by selecting highly salient patches from a set of generated image patches, employing an importance ranking mechanism. On the other hand, \textsc{DaTP} adaptively transforms learnable text tokens using supercategory names to generate subcategory-specific discrepancy descriptions, which serve as additional discriminative cues.
To maintain semantic coherence, a multimodal adaptation scheme is proposed to align the vision and text prompt into a common semantic space, enabling cross-modal collaborative reasoning through a Vision-Language Fusion Module (\textsc{VLFM}). This further improves the performance of multimodal prompting on FGVC tasks. 
However, due to the absence of specific annotations for fine-grained images and the availability of only subcategory labels or indexes, there is limited supervision for optimizing \textsc{MP-FGVC}. Thereby, a two-stage training strategy is proposed.
In the first stage, we keep the text encoder of the pre-trained CLIP fixed and optimize the image encoder with the multimodal prompts scheme. In the second stage, the image encoder, text encoder, and multimodal prompts scheme remain static. Together, they provide subcategory-specific visual embeddings and discrepancy-aware textual embeddings, facilitating the multimodal adaptation scheme to effectively exploit multimodal prompting for FGVC.

In summary, the primary contributions of this work are as follows: (1) We propose \textsc{MP-FGVC}, a novel multimodal prompting solution for fine-grained visual classification. To the best of our knowledge, this is the first approach that efficiently adapts the image-text pre-trained CLIP model to FGVC tasks in close-set scenarios. (2) Our \textsc{MP-FGVC} encompasses a multimodal prompts scheme and a multimodal adaptation scheme. The former explicitly highlights the subcategory-specific discrepancies from both the vision and language perspectives, while the latter achieves cross-modal collaborative reasoning based on the former to further enhance multimodal prompting in FGVC. (3) \textsc{MP-FGVC} employs a two-stage training strategy to fully leverage the cross-modal description ability of the CLIP model, resulting in improved FGVC with a simple modification. Experimental evaluations conducted on four FGVC datasets demonstrate the remarkable performance achieved by our proposed \textsc{MP-FGVC}.

%% file: sections/relatework.tex
\section{Related Wrok}
\subsection{Fine-Grained Visual Classification}
Existing CNN-based approaches for FGVC can generally be categorized into two groups. 
The first group, known as feature-encoding methods~\cite{YuZZZY18, GaoHWHS20,mrdmn}, focuses on learning more comprehensive features by integrating higher-order information or exploring various training strategies. 
The second group, referred to as part-localization methods~\cite{GeLY19,LiuXZMYZ20,TangYLT22}, aims to identify multiple local regions for capturing more discriminative features.
Recently, vision Transformers (ViTs)~\cite{DosovitskiyB0WZ21} have exhibited impressive performance in image classification on large-scale datasets and are currently being actively employed in FGVC~\cite{rams-trans,siamtrans,ielt}. By virtue of their patch-based processing and powerful self-attention mechanism, ViT-based methods can outperform CNN-based methods~\cite{api-net,cross-X,prnet} by solely relying on a pure Transformer encoder architecture along with diverse patch selection mechanisms.
However, these methods exclusively rely on uni-modal visual representation from a pre-defined closed set and completely disregard corresponding high-level language descriptions that encompass rich semantic information that can be generated by vision-language models. In this paper, we are the first to exploit the cross-modal description capabilities of CLIP to establish associations between subcategory-specific visual disparities in images and their corresponding textual descriptions.
This simple but effective approach yields significant improvements on FGVC.

\subsection{Vision-Language Learning}
Vision-Language (VL) models, comprising image and text encoders, are trained in a contrastive manner using large-scale image-text pairs to learn a common feature space for images and textual labels. 
Recent advancements in contrastive training with large-scale Internet data, such as CLIP~\cite{clip}, ALIGN~\cite{aligan}, and FILIP~\cite{flip}, have further scaled up VL models. 
Specifically, CLIP~\cite{clip} has achieved remarkable success in zero-shot tasks by leveraging text prompts to transfer visual concepts to downstream tasks. Consequently, numerous subsequent approaches~\cite{maple,mpmm,fgvp} have been proposed to exploit pre-trained VL models for diverse downstream tasks.
Drawing inspiration from recent developments in NLP, prompt techniques have begun to be explored in multimodal computer vision problems for transferring knowledge from large-scale VL models to downstream tasks. In the case of ViTs, some methods~\cite{fptl,vitacip} employ trainable prompts to guide pre-trained ViTs. Conversely, certain methods~\cite{coop} introduce learnable vectors into the text encoder of CLIP for transfer learning in image recognition tasks. For fine-grained understanding, some approaches propose utilizing pre-trained CLIP to address the problem of fine-grained image retrieval in open-set scenarios. However, to the best of our knowledge, no related work has specifically focused on how to develop a pre-trained CLIP model to solve FGVC tasks in close-set scenarios. The primary challenge lies in existing prompt strategies tailored towards capturing supercategory-level semantics rather than more detailed subcategory-level discrepancies. In response, we propose a multimodal prompts scheme and a multimodal adaptation scheme to enhance the sensitivity of the pre-trained CLIP model for solving FGVC tasks.

%% file: sections/preliminary.tex
\section{Preliminary}
\subsection{Review of Vision Transformer (ViT)} 
For the input image $\mathbf{I}$, ViT architecture initially split it into $N = N_{h} \times N_{w}$ non-overlapping patches of size $P\times P$. Here, $N_{h}$ and $N_{w}$ represent the number of patches in each column and row of the input image, respectively. Subsequently, these patches are transformed into embedding tokens $\mathbf{E} = \lbrack \mathbf{E}^{1}, \mathbf{E}^{2}, \cdots, \mathbf{E}^{N} \rbrack \in \mathbb{R} ^{N \times D}$ using a learnable linear projection $\mathbf{P}_{emb} \in \mathbb{R}^{P^2 \times D}$, where $D$ denotes the output dimension for each token. Lastly, the embedded tokens $\textbf{E}$ and the class token $\mathbf{E}^{class} \in \mathbb{R}^{D}$ are concatenated and combined with an additional position embedding $\mathbf{E}_{pos} \in \mathbb{R} ^{(N+1) \times D}$ to form the initial input token sequence as $\mathbf{E}_{0} = \lbrack \mathbf{E}^{class}, \mathbf{E}^{1}, \mathbf{E}^{2}, \cdots, \mathbf{E}^{N} \rbrack + \mathbf{E}_{pos}$.
The ViT encoder comprises $L$ transformer layers, each consisting of multi-head self-attention (MHSA) and multi-layer perception (MLP) blocks. Thus, when provided with the input $\mathbf{E}_{i-1}$, the output of the $i$-th layer can be expressed as:
\begin{equation}
\begin{gathered}
    \mathbf{E}_{i}^{\prime} = \textrm{MHSA}(\textrm{LN}(\mathbf{E}_{i-1})) + \mathbf{E}_{i-1}, \\
    \mathbf{E}_{i} = \textrm{MLP}(\textrm{LN}(\mathbf{E}_{i}^{\prime})) + \mathbf{E}_{i}^{\prime}.
\end{gathered}
\end{equation}
Here, $i = \{ 1, 2, \cdots, L \}$, and $\textrm{LN}(\cdot)$ denotes the layer normalization. The first token of the last layer, $\mathbf{E}_{i}^{0}$, in ViT serves as the global feature representation and is forwarded to the classifier head to generate the final predictions.

\subsection{Review of CLIP}
We provide a brief overview of the vision-language model, focusing specifically on CLIP. CLIP consists of an image encoder, denoted as $\mathcal{I(\cdot)}$, and a text encoder, denoted as $\mathcal{T(\cdot)}$. 
The image encoder $\mathcal{I(\cdot)}$ maps high-dimensional images to low-dimensional feature embeddings. The text encoder $\mathcal{T(\cdot)}$ generates text representations through an additional prompt learning scheme for a text description such as ``a photo of a [$\texttt{class}$]''. In CLIP, the placeholder $\texttt{class}$ is replaced with concrete labels corresponding to the images.
For a batch of image-text pairs, denoted as ($img_i, text_i$) where $i \in \lbrace 1, \cdots, B \rbrace$, the image encoder $\mathcal{I(\cdot)}$ and text encoder $\mathcal{T(\cdot)}$ of CLIP produce visual embeddings $\mathbf{V}_i$ and text embeddings $\mathbf{T}_i$ as $\mathbf{V}_i = \mathcal{I}(img_i)$ and $\mathbf{T}_i = \mathcal{T}(text_i)$.
The learning objective of CLIP includes an image-to-text contrastive loss, $\mathcal{L}_{i2t}$, and a text-to-image contrastive loss, $\mathcal{L}_{t2i}$, which align the learned embedding spaces for images and text, respectively. These losses are formulated as follows:
\begin{equation}\label{eq2}
\begin{gathered}
    \mathcal{L}_{i2t} = -\log\frac{\textrm{exp}(\textrm{cos}(\mathbf{V}_i, \mathbf{T}_i))}{\sum_{j=1}^{B}\textrm{exp}(\textrm{cos}(\mathbf{V}_i, \mathbf{T}_j))}, \\
    \mathcal{L}_{t2i} = -\log\frac{\textrm{exp}(\textrm{cos}(\mathbf{V}_i, \mathbf{T}_i))}{\sum_{j=1}^{B}\textrm{exp}(\textrm{cos}(\mathbf{V}_j, \mathbf{T}_i))},
\end{gathered}
\end{equation}
where $\textrm{cos}(\cdot, \cdot)$ denotes the cosine similarity functions. 

%% file: sections/Methodology.tex
\section{Methodology} \label{Methodology}
The overall pipeline of our proposed \textsc{MP-FGVC} is depicted in Fig.~\ref{fig:overview}. Obviously, our framework comprises two key components: the \textbf{multimodal prompts scheme} and the \textbf{multimodal adaptation scheme}, both of which are constructed using a pre-trained CLIP model through a two-stages training process. 
In the first stage, we refine the multimodal prompts scheme to empower the pre-trained CLIP model in recognizing subcategory-specific semantics in the vision prompt, as well as generating discrepancy-aware descriptions in the text prompt.
Subsequently, in the second stage, we optimize the multimodal adaptation scheme to fully exploit the benefits derived from the multimodal prompts scheme specifically for FGVC.

\subsection{Multimodal Prompts Scheme} 
Subtle yet discriminative discrepancies are widely acknowledged as being crucial for FGVC. However, the conventional ViTs employed in existing CLIP-like vision-language models are originally designed to represent common visual concepts and recognize them into coarse categories, rather than effectively capturing subtle differences to predict subordinate categories of a given object. 
To address this, we propose a multimodal prompts scheme that encompasses two components: {subcategory-specific vision prompt} and {discrepancy-aware text prompt}.

\subsubsection{Subcategory-specific Vision Prompt.} 
An important challenge in FGVC is accurately localizing discriminative regions that capture subcategory-specific discrepancies among confusing subcategories. 
However, the ViT treats all patch-divided image regions equally, resulting in the inclusion of much redundant patch information that can negatively impact the final decision. This limitation restricts the applicability of the pre-trained CLIP model in FGVC tasks. 
To address this limitation,  we propose a parameter-free operation called Subcategory-specific Vision Prompt (\textsc{SsVP}), which can be applied to the penultimate layer of the ViT to modify its input to the last layer. In \textsc{SsVP}, the importance scores generated by $M$ heads in MHSA are considered to select discriminative tokens that reflect the attention given to each region. Let $\mathbf{A}=[\mathbf{A}^{1}, \mathbf{A}^{2}, \cdots \mathbf{A}^{m}, \cdots, \mathbf{A}^{M}]$ denote the attention scores of the class token in the penultimate layer, where $\textbf{A}^{m} \in \mathbb{R}^{N}$ represents the attention score of the $m$-th head and reflects the attention degree of class prediction towards each region. 
As shown in Fig.~\ref{fig:overview}, we first aggregate the attention scores of all heads and then select the top-$k$ valuable tokens with high activation values as $\mathbf{id} = \texttt{TopK} (\sum_{i=1}^{M}\textbf{A}^{i}) \in \mathbb{Z}^{K}$,
where $\texttt{TopK}(\cdot)$ returns the index $\mathbf{id}$ of the first $k$ tokens based on the aggregated attention scores, sorted in descending order. Instead of treating all tokens equally, we select $k$ tokens by $\mathbf{id}$ and concatenate them along with the class token to form the input sequence of the $L$-th transformer layer, denoted as:
\begin{equation} \label{param_l}
    \mathbf{E}_{L-1} = \lbrack \mathbf{E}^{class}_{L-1}, \mathbf{E}^{\mathbf{id}(1)}_{L-1}, \mathbf{E}^{\mathbf{id}(2)}_{L-1}, \cdots, \mathbf{E}^{\mathbf{id}(k)}_{L-1} \rbrack.
\end{equation}
By replacing the original entire input sequence of the last transformer layer with tokens that solely cover the discriminative regions of an object, the proposed \textsc{SsVP} encourages the image encoder to focus more on the subtle differences between different subcategories while disregarding less discriminative regions such as background or common features, thereby facilitating fine-grained understanding.

\subsubsection{Discrepancy-aware Text Prompt.} 
The prompt strategies employed by existing CLIP-like VL models primarily focus on capturing supercategory-level semantics rather than detailed visual discrepancies necessary for distinguishing fine-grained objects. Consequently, the acquisition of subcategory-specific visual discrepancies in FGVC tasks is constrained, despite the provision of concrete subcategory names for text prompts. 
To maintain semantic coherence with the subcategory-specific vision prompt, we propose a Discrepancy-aware Text Prompt (\textsc{DaTP}) operation that supplements the lack of textual information by per-training a set of learnable text tokens. These tokens automatically generate appropriate text descriptions about more detailed visual discrepancies among subcategories. Specifically, the text prompt template fed into $\mathcal{T(\cdot)}$ is designed as $\mathbf{P}_{class}^{c}=  \lbrack \mathbf{X}_1, \mathbf{X}_2, \cdots,  \mathbf{X}_j, \cdots, \mathbf{X}_J, \texttt{class++}\rbrack$, where $c\in\{1,\cdots,C\}$ and $C$ denotes the number of subcategories. 
Each $\mathbf{X}_j\in\mathcal{R}^D$ represents the $j$-th learnable text token, and $D$ represents the dimension of the token vector. Notably, $\texttt{class++}\in \mathcal{R}^{D}$ corresponds to the word embedding generated by the \textbf{supercategory} label (\emph{e.g.,}~bird/dog/food), rather than specific subcategory labels. To ensure semantic coherence with the subcategory-specific vision prompt, each text prompt $\mathbf{P}_{class}^{c}$ serves as a discrepancy-aware textual description and shares the same supercategory name $\texttt{class++}$ within a dataset. thereby improving the computation efficiency. Ultimately, these learnable prompt vectors, in conjunction with the supercategory names, construct a discrepancy-aware text prompt that the text encoder of the CLIP model can understand. Consequently, this mechanism generates subcategory-specific discrepancy descriptions, which serve as additional discriminative cues.

\subsection{Multimodal Adaptation Scheme} 
The objective of the multimodal prompts scheme is to enhance the sensitivity of the pre-trained CLIP model to subcategory-specific discrepancies from the perspectives of both vision and text prompts. To enhance the performance of multimodal prompting on FGVC, we additionally introduce a multimodal adaptation scheme by incorporating a Vision-Language Fusion Module (\textsc{VLFM}) to align the vision and text prompts in a shared semantic space, thereby facilitating collaborative reasoning across modalities.

\subsubsection{Vision-Language Fusion Module.} As shown in Fig.~\ref{fig:overview}, VLFM first mutually aligns the vision prompt and text prompts in a semantic space using a cross-modal attention mechanism. Specifically, we consider the class token of the ViT as the subcategory-specific visual embedding, denoted as $\mathbf{E}_V$, and the textual features generated by the learned text tokens $\mathbf{P}_{class}=\{\mathbf{P}_{class}^{1}, \mathbf{P}_{class}^{2}, \cdots, \mathbf{P}_{class}^{C}\}$ as the textual embeddings that are aware of the subcategory-specific discrepancies, denoted as $\mathbf{E}_T$. 
With the multimodal input $[\textbf{E}_V, \mathbf{E}_T]$, we generate three visual embeddings, namely the query ($\mathbf{Q}_0$), key ($\mathbf{K}_0$), and value ($\mathbf{V}_0$), using a $1\times1$ convolution layer. 
Similarly, a key-value textual embedding pair ($\mathbf{K}_c-\textbf{V}_c$) is generated for each textual embedding $\mathbf{E}_{T}^{c}$ ($c \in \lbrace 1, \cdots, C\rbrace$) using the same process. 
Thus, the cross-modal attention vector $\mathbf{S}_{i}$ is computed between the visual query vector $\mathbf{Q}_0$ and all key vectors \textbf{K}$_i$~($i = \lbrace 0, 1, \cdots, C\rbrace$) as $\mathbf{S}_i = \textrm{Softmax}(\frac{\mathbf{Q}_0\mathbf{K}_i^{T}}{\sqrt{D}})$, where $D$ denotes the embedding dimension and $T$ represents the transpose operation.
Utilizing the cross-modal attention $\mathbf{S}_i$, the subcategory-specific descriptions from the text prompts are transformed into subcategory-specific visual discrepancies, given by $\mathbf{E}_{V}^{\prime} = \sum_{i=0}^{C}\mathbf{S}_i \otimes \mathbf{V}_i$, which aims to recover the missing discrepancies and form a discriminative and generalized visual embedding. 
Furthermore, to incorporate more complete discrepancies into the \textsc{VLFM}, we introduce two $1\times1$ convolution layers that constitute a translation layer denoted as $\textrm{Trans}(\cdot)$. This layer is applied to $\mathbf{E}_{V}^{\prime}$ to obtain $\hat{\textbf{E}}_{V} = \textrm{Trans}(\textbf{E}_{V}^{\prime})$. The first convolution layer, which has an expansion factor of $4$, serves as an expansion layer, while the second layer maps the feature dimension back to the original input dimension. Finally, we select the refined multimodal feature $\hat{\textbf{E}}_{V}$ for classification.

\subsection{Optimization} 
In this paper, we aim to fully exploit multimodal prompting for FGVC by utilizing the recently introduced CLIP model. However, in most FGVC tasks, only subcategory labels or indexes are available, which lack specific textual descriptions for fine-grained understanding.  As a result, the adoption of CLIP-like VL models in FGVC has been limited. To address this issue,  we propose a two-stage training strategy illustrated to enhance the sensitivity of the proposed MP-FGVC to subcategory-specific discrepancies, thereby facilitating efficient adaptation for FGVC tasks.

\subsubsection{The first training stage.} In this stage, we fix the parameters of $\mathcal{T}(\cdot)$ and optimize only the parameters of the image encoder $\mathcal{I}(\cdot)$ and the learned tokens in \textsc{DaTP} are optimized. 
Similar to CLIP, we employ the image encoder $\mathcal{I}(\cdot)$ and the text encoder $\mathcal{T}(\cdot)$ to obtain the visual embeddings $\mathbf{V}_i \in \mathbb{R}^{D}$ and the textual embeddings $\mathbf{T}_i \in \mathbb{R}^{C \times D}$. However, instead of using the original text as input to $\mathcal{T}(\cdot)$, we replace it with $\mathbf{P}_{class}$, where $\mathbf{P}_{class} = \left[ \mathbf{P}_{class}^{1}, \mathbf{P}_{class}^{2},\ldots, \mathbf{P}_{class}^{C} \right]$. Here, we use $\mathbf{T}_i^{c}$ to indicate the textual feature of subcategory $c$. To optimize $\mathcal{I}(\cdot)$, we simply utilize $\mathbf{V}_i$ to generate subcategory predictions $\hat{y}_1$ and compute the cross-entropy loss $\mathcal{L}_{v}$ as follows:
\begin{equation}\label{eq4}
    \mathcal{L}_{v} = \textrm{CrossEntropy}(\hat{y}_1,~y),
\end{equation}
where $y$ is the one-hot label. By minimizing $\mathcal{L}_{v}$, we aim to project the semantic features into a new space where subcategory-specific discrepancies are highlighted. Furthermore, to generate discrepancy-aware textual features, we employ the image-to-text contrastive loss $\mathcal{L}_{i2t}$ (defined in Eq.~\ref{eq2}) to optimize the learnable tokens in \textsc{DaTP}. This loss is computed as:
\begin{equation}
    \mathcal{L}_{i2t} = -\sum_{c=1}^C y_c \cdot \log\frac{\exp(\cos(\mathbf{V}_i, \mathbf{T}_i^{c})/\tau)}{\sum_{k=1}^{C}\exp(\cos(\mathbf{V}_i, \mathbf{T}_i^{k})/\tau)},
\end{equation}
where $\tau$ is the temperature parameter and is set to $0.07$. By minimizing both $\mathcal{L}_{v}$ and $\mathcal{L}_{i2t}$, the gradients are back-propagated through the unfixed $\mathcal{I}(\cdot)$ and fixed $\mathcal{T}(\cdot)$ to identify subcategory-specific semantics in the vision prompt and generate discrepancy-aware descriptions in the text prompt. Thus, the overall loss function of the first stage is formulated as:
\begin{equation}
    \mathcal{L}_{stage1} = \mathcal{L}_{v} + \mathcal{L}_{i2t}.
\end{equation}

\subsubsection{The second training stage.} In this stage, we fix the parameters of $\mathcal{I}(\cdot)$ and $\mathcal{T}(\cdot)$, and only optimize \textsc{VLFM} to align the vision prompt and text prompts in the semantic space, in order to fully exploit the advantages of the multimodal prompts scheme for final classification. We utilize the visual features $\hat{\textbf{E}}_{V}$ output by \textsc{VLFM} to predict the subcategory label $\hat{y}_{2}$ and calculate the cross-entropy loss $\mathcal{L}_{vlfm}$ as in Eq.~\ref{eq4}. Hence, the loss for the second training stage is:
\begin{equation}
    \mathcal{L}_{stage2} = \mathcal{L}_{vlfm}=\textrm{CrossEntropy}(\hat{y}_2,~y).
\end{equation}

%% file: sections/Experiments.tex
\section{Experiments}
\subsection{Experiments Setup}
\subsubsection{Dataset.} Experiments are conducted to evaluate the effectiveness of the proposed method on four fine-grained datasets, including
CUB-200-2011~\cite{cub}, Stanford Dogs~\cite{dog}, NABirds~\cite{nabird}, and Food101~\cite{food101}.

\textbf{CUB-200-2011} comprises $11,788$ bird images from 200 bird species, which are officially split into $5,994$ training images and $5,794$ test images.

\textbf{Stanford Dogs} consists of $20,580$ images depicting 120 dog variants, with $12,000$ images allocated for training and $8,580$ images designated for testing.

\textbf{NABirds} contains $48,562$ images showcasing North American birds across $555$ sub-categories. It is split into $23,929$ training images and $24,633$ test images.

\textbf{Food101} encompasses $101$ different kinds of foods, totaling $101,000$ images. Within each class, $250$ test images are carefully reviewed for accuracy, while the remaining $750$ training images may contain some noise.

\subsubsection{Implementation details.} To ensure a fair comparison with other methods, we employ the VIT-B-16~\cite{vit}, pre-trained on ImageNet21K, as our image encoder ~$\mathcal{I}(\cdot)$, following TransFG~\cite{transfg}. The text encoder~$\mathcal{T}(\cdot)$, is a pre-trained Transformer model from CLIP~\cite{clip}. All input images are resized to  a resolution of $448 \times 448$. In the first training stage, we initialize the learning rate as $3$e-$2$, except for Stanford Dogs where it is initialized as $3$e-$3$. 
The number of training epochs is set to $30$ for both CUB-200-2011 and Stanford Dogs, while the remaining datasets are trained for $10$ epochs. 
For the second training stage, we initialize the learning rate as 1e-3, except for Stanford Dogs where it is initialized as 1e-4. The number of training epochs is set to $10$ for both CUB-200-2011 and Stanford Dogs, while the other datasets are trained for $3$ epochs. 
Both training stages utilize the SGD optimizer and employ cosine annealing as the optimization scheduler. The batch size is fixed at $32$.

\begin{table}[tb!]
    \centering
    \resizebox{0.95\linewidth}{!}{
    \begin{tabular}{c|c|c|c}
        \toprule
        Method &Backbone &CUB & Dogs \\
         \midrule
        MRDMN~\cite{mrdmn} &CNN &88.8 &89.1 \\
        FDL~\cite{fdl} &CNN &89.1 &84.9 \\
        MSHQP~\cite{mshqp} &CNN &89.0 &90.4 \\
        API-Net~\cite{api-net} &CNN &90.0 & 90.3 \\
       
        PRIS~\cite{pris} &CNN &90.0 &\underline{90.7} \\
        CAL~\cite{cal} &CNN &90.6 &88.7 \\
        \midrule
         TransFG*~\cite{transfg} &ViT-B-16 &91.1 &90.5 \\
         IELT*~\cite{ielt} &ViT-B-16 &90.9 &90.6 \\
         SIM-Trans*~\cite{siamtrans} &ViT-B-16 &\underline{91.3} &89.2 \\
         MP-FGVC (Ours) &ViT-B-16 &\textbf{91.8} &\textbf{91.0} \\
        \bottomrule
    \end{tabular}}\vspace{-2mm}
    \caption{\small Comparisons with competitive methods on CUB-200-2011 and Stanford Dogs, * indicates the result reproduced according to the source code.}
    \label{tab:cub_dogs}\vspace{-4mm}
\end{table}

\begin{table}[tb!]
    \centering
    \resizebox{0.95\linewidth}{!}{
    \begin{tabular}{c|c|c}
        \toprule
        Method &Backbone &NABirds \\
        \midrule
        Cross-X~\cite{cross-X} &CNN &86.4  \\
        DSTL~\cite{dstl} &CNN &87.9 \\
        GHRD~\cite{ghrd} &CNN &88.0\\
        API-Net~\cite{api-net} &CNN &88.1 \\
        PRIS~\cite{pris} &CNN &88.4\\
        MGE-CNN\cite{mge-cnn} &CNN &88.6 \\
       \midrule
         TransFG*~\cite{transfg} &ViT-B-16 &\underline{90.5}\\
         IELT*~\cite{ielt} &ViT-B-16 &90.1\\
         SIM-Trans*~\cite{siamtrans} &ViT-B-16 &90.3 \\
         MP-FGVC (Ours) &ViT-B-16 &\textbf{91.0}\\
        \bottomrule
    \end{tabular}}\vspace{-2mm}
    \caption{\small Comparisons with competitive methods on NABirds.}
    \label{tab:nabirds}\vspace{-4mm}
\end{table}

\begin{table}[tb!]
    \centering
    \resizebox{0.95\linewidth}{!}{
    \begin{tabular}{c|c|c}
        \toprule
        Method &Backbone &Food101 \\
        \midrule
        DCL~\cite{DCL} &CNN &88.9  \\
        SGLANet~\cite{SGLANet} &CNN &89.7 \\
        IG-CMAN~\cite{IG-CMAN} &CNN &90.4\\
        MSMVFA~\cite{msmvfa} &CNN &90.6\\
        PRENet~\cite{prnet} &CNN &91.1 \\
       \midrule
         TransFG*~\cite{transfg} &ViT-B-16 &\underline{92.1}\\
         IELT*~\cite{ielt} &ViT-B-16 &91.4\\
         SIM-Trans*~\cite{siamtrans} &ViT-B-16 &91.9 \\
         MP-FGVC (Ours) &ViT-B-16 &\textbf{93.0}\\
        \bottomrule
    \end{tabular}}\vspace{-2mm}
    \caption{\small Comparisons with competitive methods on Food101.}
    \label{tab:food101}\vspace{-4mm}
\end{table}

\subsection{Comparison with State-of-the-Art Methods}
The comparison results for CUB-200-2011 and Stanford Dogs datasets are presented in Table ~\ref{tab:cub_dogs}, while the results for the NABirds dataset can be found in Table~\ref{tab:nabirds}. 
Furthermore, the comparison results for the Food101 dataset are provided in Table~\ref{tab:food101}. From these tables, it can be observed that our proposed method  outperforms other state-of-the-art methods across all datasets.
Specifically, our method achieves an improvement of $0.5\%$, $0.3\%$, $0.5\%$, and $0.9\%$ in terms of the Top-1 Accuracy metric compared to the second-best results on the CUB-200-2011, Stanford Dogs, NABirds, and Food101 datasets, respectively. 
Notably, our method exhibits substantial improvements over the ViT-based methods, which solely rely on visual training. More precisely, our method attains improvements of $0.5\%$-$0.7\%$, $0.4\%$-$1.8\%$, $0.5\%$-$0.9\%$, and $0.9\%$-$1.6\%$ on the CUB-200-2011, Stanford Dogs, NABirds, and Food101 datasets, respectively. These findings serve as compelling evidence of the effectiveness of our proposed \textsc{MP-FGVC}.

\subsection{Ablation Studies and Analysis} 
\subsubsection{Efficacy of various components.} The proposed \textsc{MP-FGVC} comprises three essential components: \textsc{SsVP} and \textsc{DaTP} from multimodal prompts scheme, and \textsc{VLFM} from multimodal adaptation scheme. We conducted ablation experiments on these components, and the results are reported in Table~\ref{tab:ablation}, with the Baseline representing the pure ViT.
The introduction of \textsc{DaTP} led to a $0.5\%$ improvement in multimodal prompting performance on the CUB-200-2011 dataset and a $0.4\%$ improvement on the NABirds dataset.  
By utilizing \textsc{VLFM} to enhance FGVC with language modality, further performance gains were achieved on both CUB-200-2011 and NABirds datasets.
This observation indicates that \textsc{DaTP} incorporates subcategory-specific discrepancy descriptions that complement the image content, thereby improving the performance. Moreover, the inclusion of \textsc{SsVP} in all model variations improved their performance on the CUB-200-2011 and NABirds datasets, demonstrating the effectiveness of selecting discriminative regions to identify subcategory-specific discrepancies. 

\begin{table}[t!]
    \centering
    \resizebox{0.95\linewidth}{!}{
    \begin{tabular}{l|c|c}
    \toprule
        Text Prompt Templates &CUB &NABirds  \\
         \midrule
         ``a photo of a \texttt{class}"&91.5 &90.7 \\
         ``a photo of $\mathbf{X}_1$, $\cdots$,  $\mathbf{X}_J$, \texttt{class}" &91.7 &90.9 \\
         ``a photo of $\mathbf{X}_1$, $\cdots$,  $\mathbf{X}_J$, \texttt{class++}" &\bf91.8 &\bf91.0 \\
         ``$\mathbf{X}_1$, $\cdots$,  $\mathbf{X}_J$, \texttt{class++}" &\bf 91.8 & \bf91.0 \\
    \bottomrule
    \end{tabular}}\vspace{-2mm}
    \caption{\small Comparisons with different text prompt design of accuracy~(\%) on CUB-200-2011 and NABirds.}
    \label{tab:text_desgin}\vspace{-4mm}
\end{table}

\begin{table}[t!]
    \centering
     \resizebox{0.95\linewidth}{!}{
    \begin{tabular}{l|c|c}
    \toprule
        Method &CUB &NABirds  \\
        \midrule
         Baseline &90.8 &90.0  \\
         Baseline + \textsc{DaTP} &91.3 & 90.4 \\
         Baseline + \textsc{DaTP} + VLFM &91.5 & 90.6  \\
         Baseline + \textsc{SsVP} &91.2 &90.3  \\
         Baseline + \textsc{SsVP} + \textsc{DaTP} &91.5 &90.8  \\
         Baseline + \textsc{SsVP} + \textsc{DaTP} + VLFM &\textbf{91.8} &\textbf{91.0}  \\
         \bottomrule
    \end{tabular}}\vspace{-2mm}
    \caption{\small The accuracy~(\%) results of component ablation study on CUB-200-2011 and NABirds.}
    \label{tab:ablation}\vspace{-6mm}
\end{table}

\subsubsection{Comparison of different vision prompt schemes.} To demonstrate the effectiveness of the proposed \textsc{SsVP}, we conduct comparative experiments with other token selection methods in existing FGVC solutions, namely, PSM~\cite{transfg} and MHVM~\cite{ielt}. The corresponding results are all listed in Table~\ref{tab:tsm_compare}. The experimental results show that our proposed \textsc{SsVP} achieves superior performance in capturing subcategory-specific discrepancies, leading to a consistent improvement in decision making.

\begin{table}[!t]
\centering%
\begin{minipage}[b]{0.22\textwidth}
\resizebox{\linewidth}{!}{
    \begin{tabular}{l|c|c}
    \toprule
         Settings &CUB &NABirds  \\
         \midrule
         Ours + \textsc{PSM} &91.4 &90.7 \\
         Ours + \textsc{MHVM} &91.2 &90.6 \\
         Ours + \textsc{SsVP} &\textbf{91.8} &\textbf{91.0} \\
         \bottomrule
    \end{tabular}}\vspace{-2mm}
    \caption{\small Analyses of different vision prompt methods on CUB-200-2011 and NABirds.}
    \label{tab:tsm_compare}
\end{minipage}%
\hspace{3mm}%
\begin{minipage}[b]{0.23\textwidth}
\centering%
\resizebox{\linewidth}{!}{
\begin{tabular}{l|c|c}
    \toprule
         Strategies &CUB &NABirds  \\
        \midrule
         One stage &91.1 &90.3 \\
         Two stage&\textbf{91.8} &\textbf{91.0} \\
    \bottomrule
    \end{tabular}}\vspace{-2mm}
    \caption{\small Comparison of different training strategies on CUB-200-2011 and NABirds.}
    \label{tab:train_strategies}
\end{minipage}\vspace{-2mm}
\end{table}

\subsubsection{Comparison of various text prompt templates.}~We conduct a comparative experiment to analyze the impact of different text prompt templates in \textsc{DaTP}. Specifically, \texttt{class} denotes the concrete subcategory name and \texttt{class++} denotes the ambiguous supercategory name. The results in Table~\ref{tab:text_desgin} demonstrate that learnable text tokens for the text prompt outperform handcrafted prompt templates such as "a photo of a \texttt{class}". This indicates that our \textsc{DaTP} can learn a set of words that describe the subcategory-specific discrepancies between subcategories, leading to improved performance on FGVC. Moreover, the performance is not significantly affected by whether "a photo of a" is used as a prefix or if the subcategory name is used as a suffix.

\begin{figure}[!t]
\centering%
\begin{minipage}[b]{0.23\textwidth}
\centering%
\includegraphics[width=\linewidth]{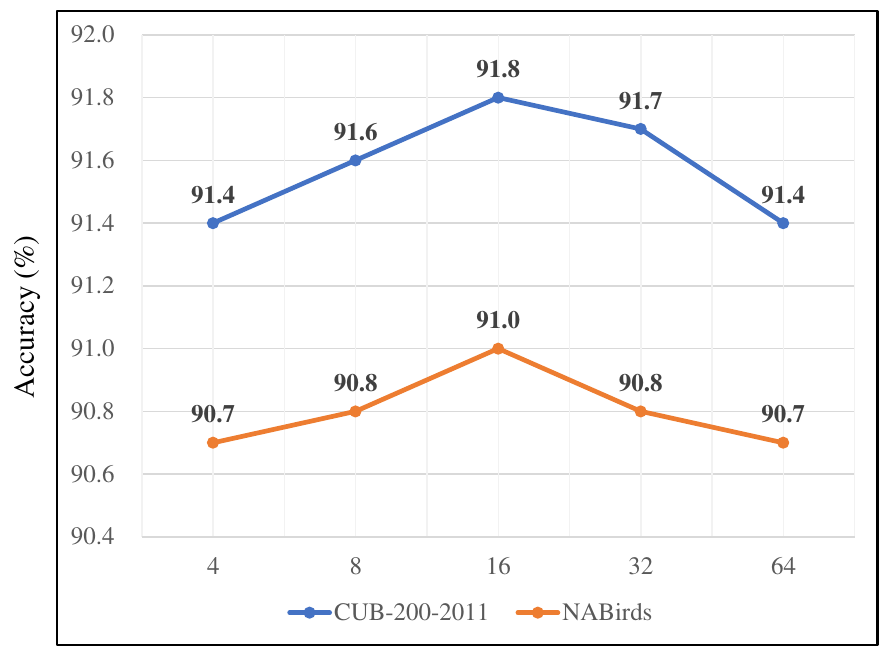}\vspace{-2mm}
\caption{\small Analyses of hyper-parameter $J$ on CUB-200-2011 and NABirds.}
\label{fig:text_length}
\end{minipage}%
\hspace{2mm}%
\begin{minipage}[b]{0.23\textwidth}
\centering%
\includegraphics[width=\linewidth]{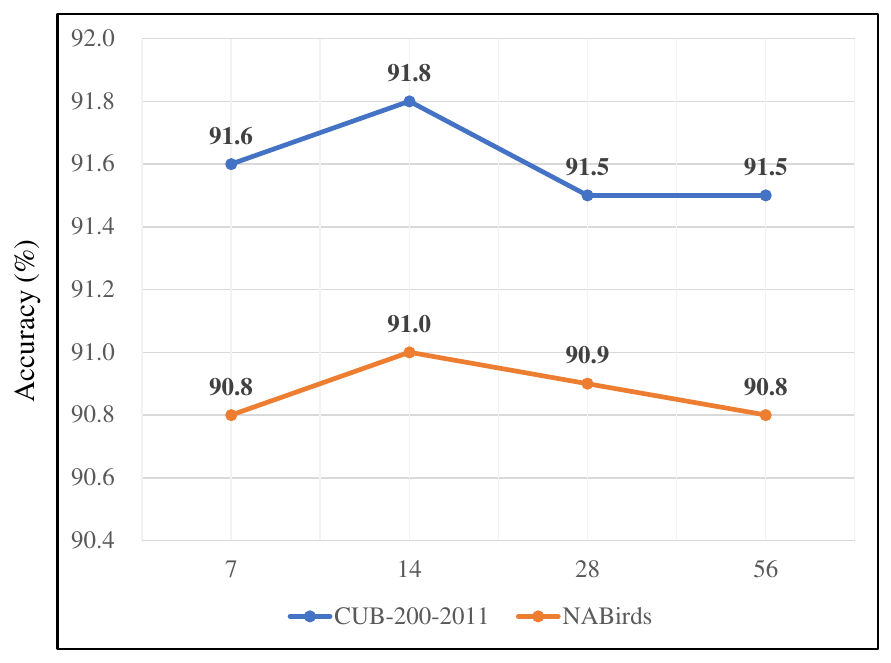}\vspace{-2mm}
\caption{\small Analyses of hyper-parameter $k$ on CUB-200-2011 and NABirds.}
\label{fig:select_number}
\end{minipage}
\vspace{-1cm}
\end{figure}

\subsubsection{Number of selected visual tokens in \textsc{SsVP}.} The performance comparison of different values of $k$, representing the number of selected visual tokens in \textsc{SsVP}, is presented in Figure~\ref{fig:select_number}. From the analysis, it is observed that  performance degrades when $k$ exceeds $14$. This performance drop may be attributed to the fact that as the number of selected tokens increases, the vision prompt may highlight the locations of objects or parts instead of the subcategory-specific differences, rendering it ineffective for FGVC. Consequently, we have chosen to set $k$ to $14$ for all datasets.

\subsubsection{Number of learnable text tokens in \textsc{DaTP}.} What is the appropriate number of learnable tokens in \textsc{DaTP} for accurately summarizing subcategory-specific discrepancies?  We conduct analyses on the length $J$ of the text prompt template, and present evaluation results in Fig.~\ref{fig:text_length}. The performance gradually increases with the number of learnable tokens but decreases when the prompt length exceeds $32$. One potential explanation for this observation is that as the length of the text prompt reaches $32$, the model tends to prioritize optimizing \textsc{DaTP} itself, hindering the optimization process for the subcategory-specific vision prompt. Thus, we set the token number in \textsc{DaTP} to $16$ for all datasets.

\subsubsection{Necessity of two-stage training strategy.} Our proposed MP-FGVC employs a two-stage training strategy to align and fuse embeddings from text and image domains. It is worth noting that integrating the learning of vision prompt, text prompt, and multimodal adaptation into single-stage training, similar to CLIP-like VL models, is possible. To determine the effectiveness of each strategy, we conduct comparison experiments between the two-stage training and the one-stage training. Table~\ref{tab:train_strategies} demonstrates that the one-stage training strategy performs less effectively. The inadequate description of subcategory-specific discrepancies by the randomly initialized text tokens during the early training stage hinders the overall optimization of the model. Therefore, we adopt the two-stage training to retain the hidden states of the pre-trained image encoder and text encoder, enabling CLIP to maintain its advantages on FGVC.

\begin{figure}[t!]
    \centering
    \includegraphics[width=0.9\linewidth]{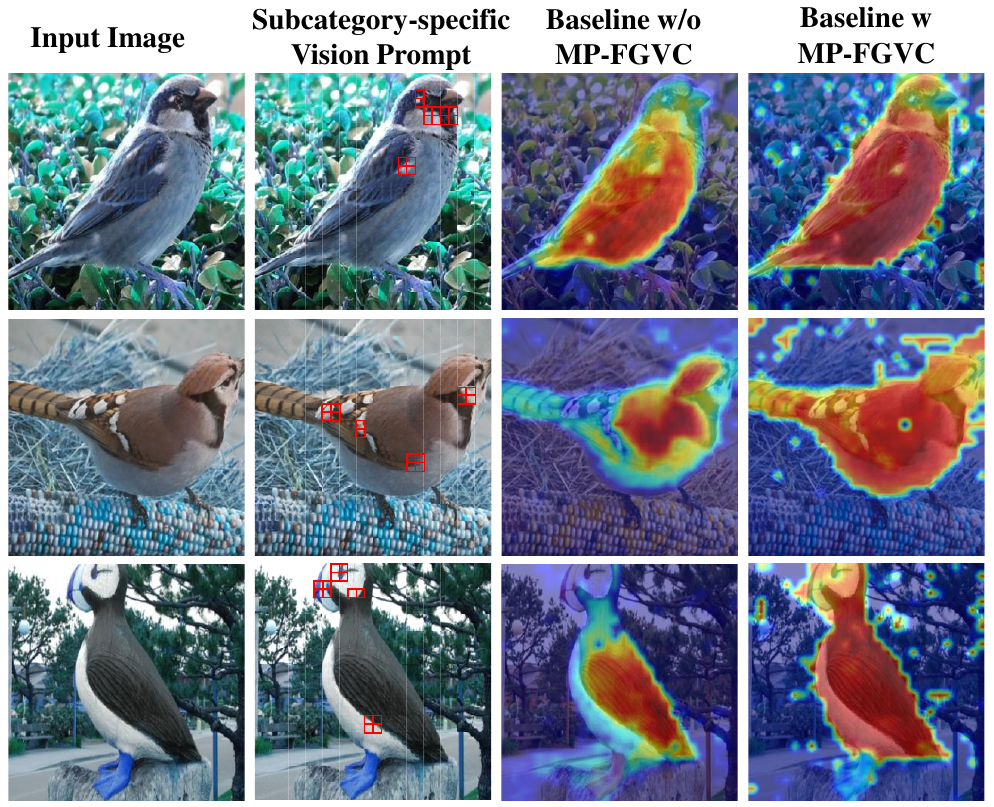}
    \caption{\small Visualization of \textsc{MP-FGVC} on CUB-200-2011.
    }
    \label{fig:visual}
    \vspace{-6mm}
\end{figure}

\subsubsection{Visualization of MP-FGVC.} We conduct visualization experiments to illustrate the effectiveness of MP-FGVC, as shown in Fig.~\ref{fig:visual}. The original input images are depicted in the first column, followed by the second column which represents the selected vision tokens ($k$=$14$) in our\textsc{ SsVP}. The third and fourth columns display the class activation maps computed using GradCAM for the baseline model with and without our proposed MP-FGVC, respectively. The visualization results demonstrate that MP-FGVC enhances the baseline model's ability to focus on a more comprehensive region, particularly in capturing more detailed discrepancies specific to subcategories, such as the head, back, abdomen, and wings in discriminative regions with color mutations.

%% file: sections/Conlusion.tex
\section{Conclusion}
In this paper, we introduce \textsc{MP-FGVC}, a novel multimodal prompting solution designed to improve fine-grained visual classification through the utilization of the cross-modal capabilities of the pre-trained CLIP model. By employing a multimodal prompts scheme and a multimodal adaptation scheme, \textsc{MP-FGVC} effectively highlights subcategory-specific discrepancies and achieves cross-modal reasoning, ultimately leading to improved FGVC performance with minimal modifications. Experimental evaluations conducted on four FGVC datasets validate the effectiveness of \textsc{MP-FGVC}. As the pioneer effort to extend the applicability of large-scale vision-language models to FGVC, \textsc{MP-FGVC} incorporates novel prompting and adaptation techniques. It is our aspiration that our findings serve as inspiration and facilitate further research on efficient adaptation methods of large-scale vision-language models for FGVC tasks.